\pdfoutput=1

\documentclass[11pt]{article}

\usepackage[]{acl}
\usepackage{times}
\usepackage{latexsym}
\usepackage{graphicx}
\usepackage[T1]{fontenc}

\usepackage[utf8]{inputenc}
\usepackage{booktabs}
\usepackage{microtype}
\usepackage{url}
\usepackage{floatrow}
\title{HumanRankEval: Automatic Evaluation of LMs\\ as Conversational Assistants}

\author{Milan Gritta, Gerasimos Lampouras, Ignacio Iacobacci\\
	Huawei Noah’s Ark Lab, London, UK\\
	\texttt{\{milan.gritta,gerasimos.lampouras,ignacio.iacobacci\}@huawei.com}
}

\begin{document}
\maketitle
\begin{abstract}

Language models (LMs) as conversational assistants recently became popular tools that help people accomplish a variety of tasks. These typically result from adapting LMs pretrained on general domain text sequences through further instruction-tuning and possibly preference optimisation methods. The evaluation of such LMs would ideally be performed using human judgement, however, this is not scalable. On the other hand, automatic evaluation featuring auxiliary LMs as judges and/or knowledge-based tasks is scalable but struggles with assessing conversational ability and adherence to instructions. To help accelerate the development of LMs as conversational assistants, we propose a novel automatic evaluation task: HumanRankEval (HRE). It consists of a large-scale, diverse and high-quality set of questions, each with several answers authored and scored by humans. To perform evaluation, HRE ranks these answers based on their log-likelihood under the LM's distribution, and subsequently calculates their correlation with the corresponding human rankings. We support HRE's efficacy by investigating how efficiently it separates pretrained and instruction-tuned LMs of various sizes. We show that HRE correlates well with human judgements and is particularly responsive to model changes following instruction-tuning. 

\end{abstract}

\begin{figure}[t]
\includegraphics[width=\linewidth]{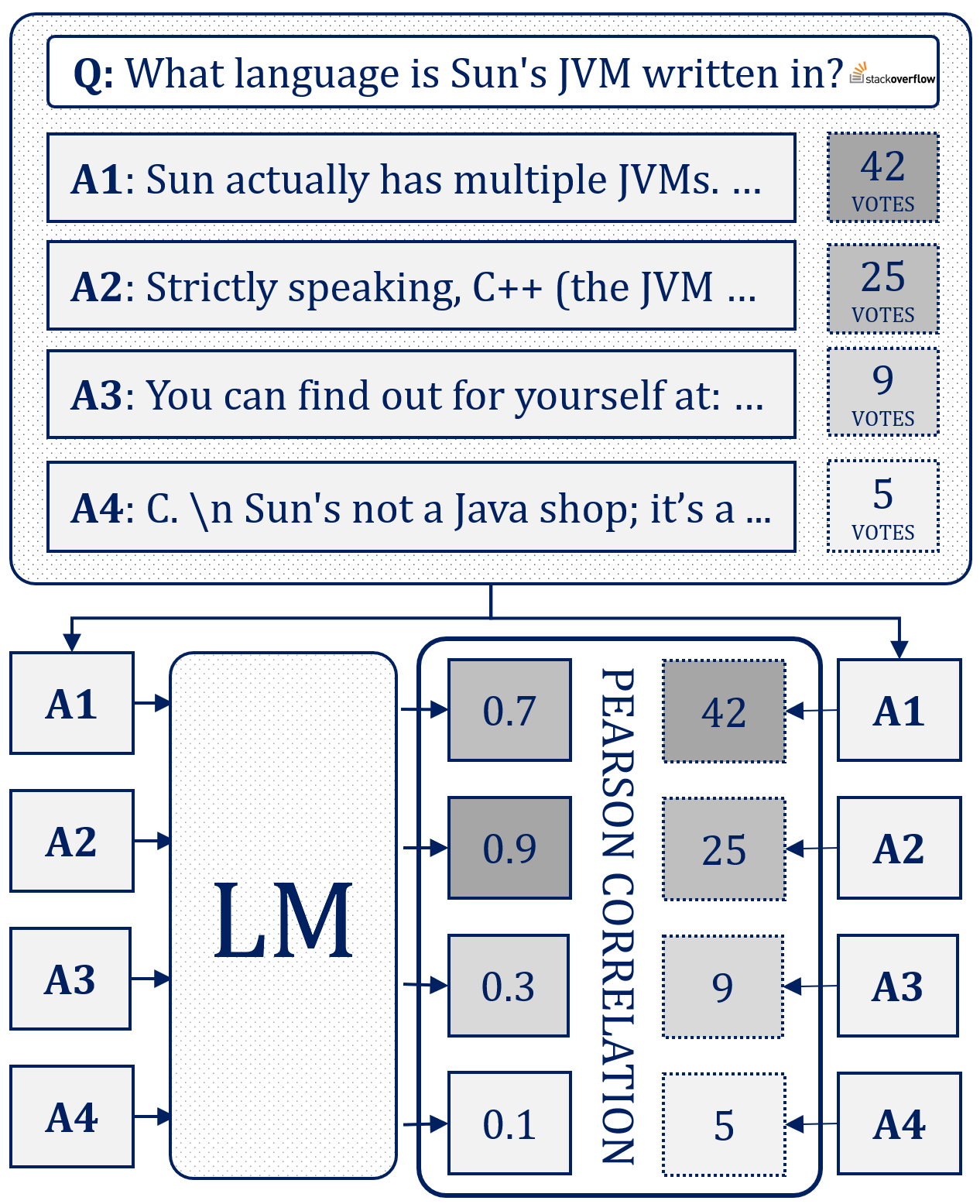}
\caption{Overview of HumanRankEval: given a question with multiple answers, we correlate human scores of each answer with the log-likelihoods of the LM. The unabridged answers can be found in Figure \ref{fig:dialogue}.}
\label{fig:headline}
\end{figure}

\begin{table*}[t]
    \centering
    \begin{tabular}{lcc}
    \toprule
    Evaluation Type        & Ground Truth                  & Metric(s)         \\\midrule
    Human                  & Human judgement    & Elo, Win-Rate, Other         \\ 
    Knowledge-based        & Human-authored text (e.g. exams, tests)      & Accuracy-like \\ 
    LM-as-a-judge          & LMs (e.g. ChatGPT, GPT-4)    & Elo, Win Rate, Other     \\ 
    \textbf{HumanRankEval} & \textbf{Human-authored text (Ranked QA pairs)} & \textbf{Correlation}  \\\bottomrule
    \end{tabular}
    \caption{Human evaluation versus relevant automatic evaluation types and their key features.}
    \label{tab:related_summary}
    \vspace{-.1mm}
\end{table*}

\section{Introduction}

The evaluation of Language Models (LMs) is a challenging problem and a prolific research subject. Many benchmarks have recently been proposed aiming to evaluate the general capabilities of LMs, covering both automatic and human evaluation \cite{chang2023survey}. Evaluating LMs' capabilities as conversational assistants, i.e. its adherence to human instructions, is particularly challenging as model inputs and outputs are more unstructured and open-ended. Ideally, human judgement should be employed to evaluate such open-ended outputs, typically either through interactive conversation with LMs \cite{zheng2023judging} or by presenting participants with outputs from different LMs and collecting their preferences \cite{ChrisvanDerLee2021}. As this approach is time-consuming and does not scale well, previous work proposed to substitute human judgement with auxiliary large LMs. However, these efforts have so far only been applied with proprietary models, e.g. GPT-4 \cite{zheng2023judging,dubois2023alpacafarm}, and with mixed results \cite{chiang2023can}. On the other hand, conventional automatic evaluation of LMs on knowledge-based tasks such as multiple-choice question-answering (QA) \cite{zellers2019hellaswag,clark2018think,lin2021truthfulqa}, can measure specific task performance in a scalable manner, but is not necessarily indicative of how an LM would perform these tasks in an open-ended conversational setting \cite{tunstall2023zephyr}. 

To this end, we introduce HumanRankEval (HRE), an automatic evaluation task for LMs as conversational assistants that comprises a novel dataset and metric. The core idea behind HRE is to measure an LM's alignment with human preferences (HP). Intuitively, given a question ($Q$) with multiple available responses ($A1, \ldots, A4$), HRE measures how well an LM's ``preference'' ranking over those answers aligns with those of humans (see Figure~\ref{fig:headline}). We approximate HP by collecting a set of questions and rated answers from StackOverflow and StackExchange. The HRE dataset covers a diverse collection of 14 topics, each containing 500 information-seeking questions paired with the top-4 answers rated (on average) by 100+ domain experts. To estimate the ``preferences'' of an LM, we obtain the log-likelihood of each answer under the model's distribution. The HRE metric is calculated as the correlation of the LM's rankings against the corresponding human rankings. We should note that we do not consider HRE as a replacement for human judgement, but rather propose its usage for fast iterations during development.

We support HRE's efficacy by investigating how effectively it separates pretrained and instruction-tuned LMs of various sizes. We then compare our results against those of other evaluation frameworks, showing that HRE correlates well with human evaluation of LMs and provides unique insights. Specifically, relative to OpenLLM, a highly popular automatic evaluation leaderboard \cite{open-llm-leaderboard}, HRE is able to more effectively differentiate pretrained and instruction-tuned LMs. Our contributions are threefold: 1) we create a large-scale, high-quality, diverse QA dataset to capture/approximate human preferences, 2) introduce an efficient automatic method to evaluate LMs as conversational assistants by measuring the correlation of LM and human preferences, and 3) perform analysis that shows HRE correlates well with human judgement and provides unique insights.\footnote{The dataset and code can be downloaded from \url{https://github.com/huawei-noah/noah-research/tree/master/NLP/HumanRankEval}.}

\section{Related Work}
\label{sec:related}

The evaluation of LMs is a highly active research topic, exemplified by a recent survey \cite{chang2023survey} that tracks over 250 papers, with over 100 of those published in just the last 12 months. There are additional surveys focused on alignment \cite{wang2023aligning}, trustworthiness \cite{liu2023trustworthy}, morals \cite{scherrer2023evaluating} and fairness \cite{li2023survey} as well as multiple benchmarks with leaderboards covering a wide variety of LM behaviours \cite{zhong2023agieval,wang2023scibench,srivastava2022beyond,chia2023instructeval,ye2023flask,liang2022holistic,dubois2023alpacafarm,liu2023agentbench,yuan2023evaluating,sun2023evaluating,ziyu-etal-2023-lens}, to list just a few. Therefore, we focus on methods relevant to evaluating LMs as conversational assistants to differentiate from prior work.

\begin{figure*}[t]
\includegraphics[width=\linewidth]{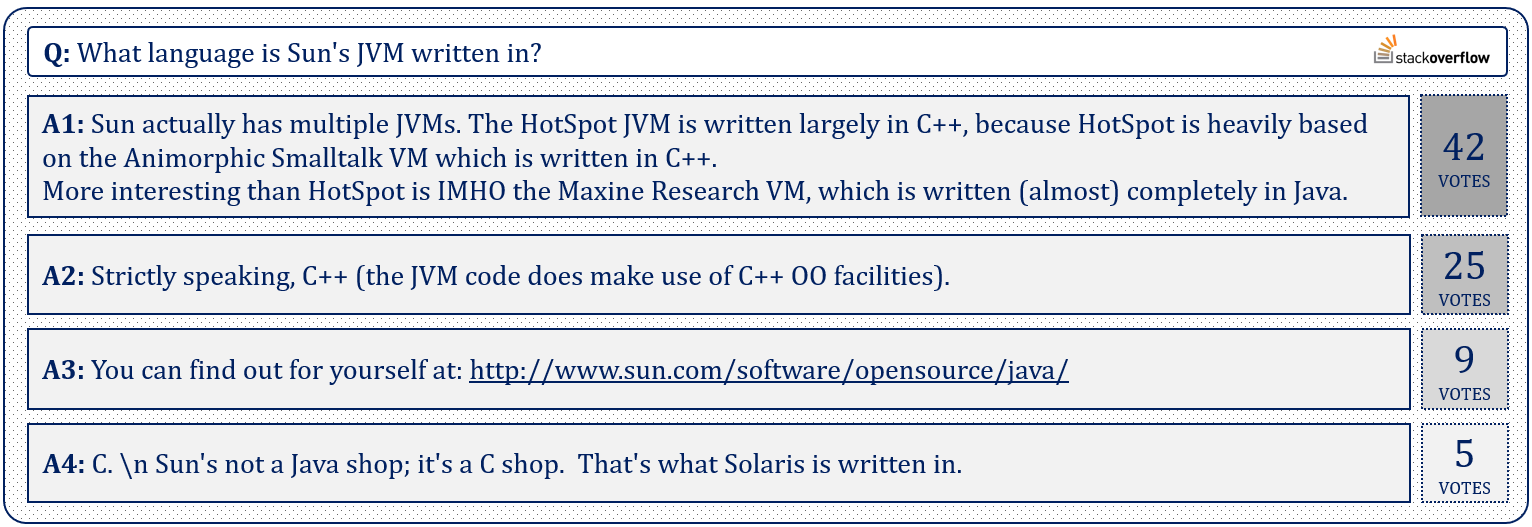}
\caption{HumanRankEval example from StackOverflow (Java topic).}
\label{fig:dialogue}
\end{figure*}

\subsection{Human Evaluation}

Due to the open-ended nature of the output, human judgement is considered the gold standard for evaluating LMs as conversational assistants \cite{ji2023beavertails,song2023preference,rafailov2023direct}, however, such evaluation is costly and can be biased \cite{wu2023style}. These issues are more prevalent in crowd-sourcing settings where participants need to be vetted to ensure their expertise and reliability, especially given that the motivations at play (e.g. to complete as many assessments as fast as possible) may run counter to the purposes of the evaluation \cite{ChrisvanDerLee2021}. Evaluation is often set up as an interactive dialog with each LM where participants are asked to rate its performance in various metrics \cite{ChrisvanDerLee2021, ji-etal-2022-achieving} or by contrasting multiple LM outputs (produced by the same input/prompt) and voting for the one that is preferred \cite{bai2022training}. The latter preferences can be converted into Elo ratings to obtain LM rankings \cite{zheng2023judging,wu2023style}. A public leaderboard that maintains such rankings is Chatbot Arena.\footnote{\label{footnote_chatbot_arena}\url{https://huggingface.co/spaces/lmsys/chatbot-arena-leaderboard}} Its game-like environment encourages users to guess the identity of two LMs at the end of an anonymous interaction. These multi-turn conversations are unstructured and depend on the interests of participants.

\subsection{Automatic Evaluation}

\subsubsection{Knowledge-based Evaluation}

A subset of automatic evaluation focuses on knowledge-based tasks with strictly-defined inputs and outputs, to enable the easy application of automatic metrics and measure performance. This is in contrast to how conversation assistants operate, where input and output is more open-ended. For LMs as conversational assistants, the focus of knowledge-based evaluation is to measure the general capabilities of the model, rather than particular performance on downstream tasks. As such, evaluation is usually applied through zero-shot or few-shot/prompt settings, without fine-tuning LMs on task-specific data. Examples include multiple-choice QA \cite{logiqa}, code generation \cite{chen2021evaluating}, Tool/API usage \cite{liu2023agentbench}, general and advanced knowledge tests \cite{hendrycks2020measuring,logiqa,cobbe2021training,zellers2019hellaswag,clark2018think,lin2021truthfulqa}, complex logical reasoning \cite{cobbe2021training}, school admission tests \cite{zhong2023agieval} and fine-grained "skill sets" evaluation \cite{ye2023flask}. Individual benchmarks are often aggregated into high-profile public rankings such as the OpenLLM Leaderboard, which we reference throughout.\footnote{\url{https://hf.co/spaces/HuggingFaceH4/open_LM_leaderboard}} Importantly, such tasks and metrics do not accurately estimate how an LM may perform on them within a conversational context, as they were not designed for this purpose. 

\subsubsection{LM-as-a-judge}

A faster alternative to human evaluation has been proposed recently, i.e. to use LMs as judges (typically larger than the LMs being judged). The most popular examples include MT-Bench \cite{zheng2023judging} and AlpacaEval\footnote{\url{https://tatsu-lab.github.io/alpaca_eval/}} \cite{dubois2023alpacafarm}. MT-Bench prompts GPT-4 to score the quality of the candidate LM on a 10-point scale over 80 two-turn conversations. AlpacaEval instructs GPT-4 to vote whether the output of the candidate LM or ChatGPT is better, resulting in a win-rate \% against GPT-3.5, using 805 manually selected prompts. However, these models are known to have biases \cite{wu2023style} and their appropriateness for LM evaluation is frequently being questioned \cite{aiyappa2023can,chiang2023can,li2023prd}. At the time of this writing, such approaches have only been explored in connection with proprietary LMs, with concerns about data privacy, frequent model (judge) changes, deprecated APIs and their associated costs.

\subsubsection{A Note on Multi-Turn Evaluation}

Even though the goal is to evaluate LMs as conversational assistants, most automatic evaluation methods (including HRE) are limited to evaluating single turn conversations. This is due to the difficulty of integrating LM interaction within an automatic task. MT-Bench contains two-turn prompts, but assumes no interaction either, with the second-turn prompt attending on a reference answer.

\section{HumanRankEval}
\label{sec:humanrankeval}

We now introduce HumanRankEval, an automatic evaluation task (comprising a novel dataset and metric) for LMs as conversational assistants. As mentioned earlier, the core idea behind HRE is to evaluate LMs by observing how an LM's ``preference'' ranking (derived from the model's log-probabilities over several answers) aligns with human-obtained rankings. To achieve this, we gather open-ended, information-seeking questions from popular online communities to capture HP. Each question comes with several answers ranked by domain enthusiasts (see Figure~\ref{fig:dialogue} for an example), indicating the order of responses (most to least preferable). Our data sources consist of StackExchange and StackOverflow. As both contain a plethora of topics, some of which may be considered subjective, we endeavoured to select the more objective/quantitative topics that we would expect to have a high degree of consensus among users, i.e. most people would agree on "good" answers. 

\subsection{StackExchange}
\label{sec:stackexchange}

StackExchange is a trusted site for communities of experts answering questions on various subjects. The data dumps were sourced from the Internet Archive\footnote{\url{https://archive.org/download/stackexchange}} and processed with Eleuther's scripts.\footnote{\url{https://github.com/EleutherAI/stackexchange-dataset}} Due to limited data availability (after filtering for quality), we set the number of questions to 500 for each topic for a uniform distribution over all domains. We selected questions from popular discussion topics: Unix-based OS, English Language, Physics, LaTeX, Software Engineering, Maths and Statistics. We also created three "mixed topics" (500 questions each) from somewhat less popular subsets that did not individually yield enough questions after filtering: \textit{CS+DB} (CodeReview, Computer Science, Data Science and Databases), \textit{App+Andr} (Apple and Android) and \textit{Lang+Sci} (Latin, Chinese, French, German, Japanese, Spanish plus Engineering, Chemistry, Biology, Earth Science and Astronomy).

\subsection{StackOverflow}
\label{sec:stackoverflow}

StackOverflow is a highly popular website and a leading community of people who contribute their expertise on a plethora of technical topics. In order to prevent HRE from being dominated by programming languages, i.e provide a balance against the more general topics of StackExchange, we selected questions from each of the following popular topics: Python, Java, HTML (includes CSS, JavaScript) and C++. The dataset was contributed by \citet{li2023textbooks}.\footnote{\url{https://huggingface.co/datasets/suriyagunasekar/stackoverflow-with-meta-data}} Once again, we set the number of questions to 500 per topic for a balanced dataset.

\begin{figure}[t]
\includegraphics[width=0.95\linewidth]{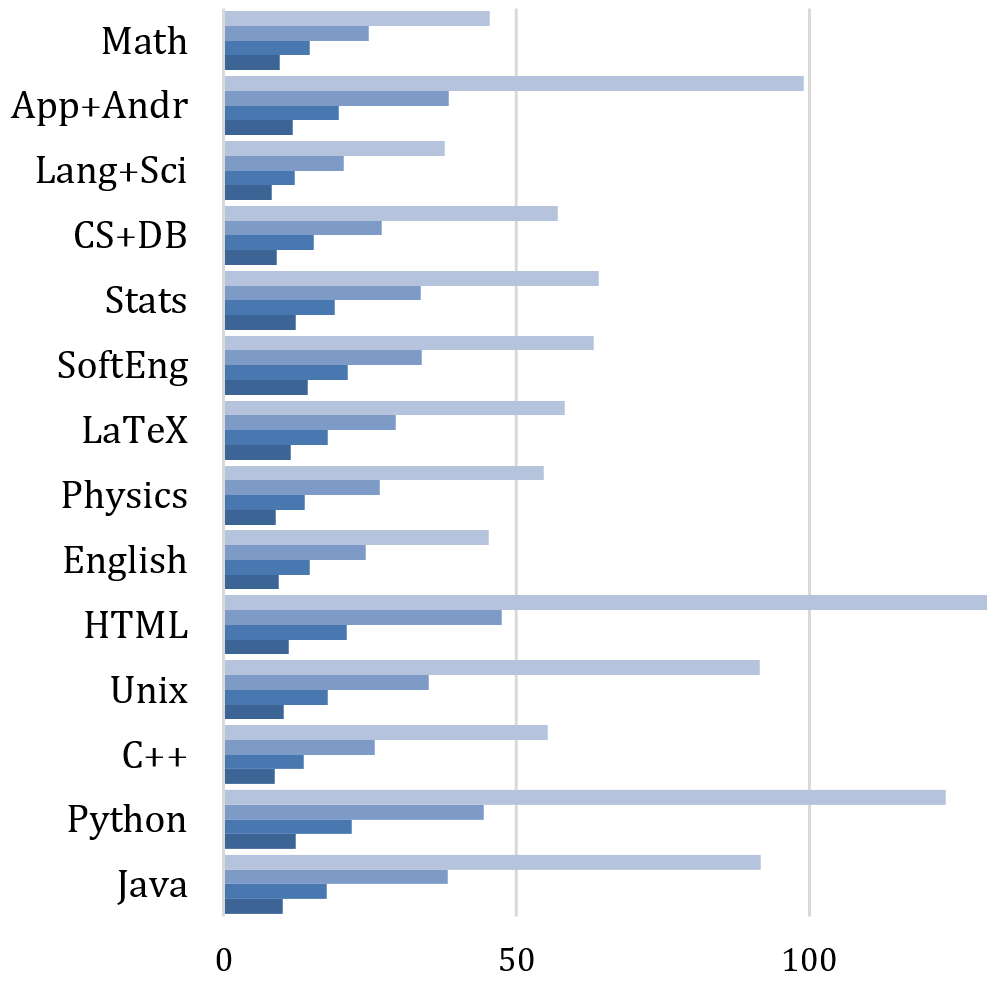}
\caption{Average votes per answer/topic. Each answer has approximately double the votes of the next answer. More details can be found in Figure \ref{fig:appendix_votes} in the Appendix.}
\label{fig:stats}
\end{figure}

\begin{figure*}[t]
\includegraphics[width=\linewidth]{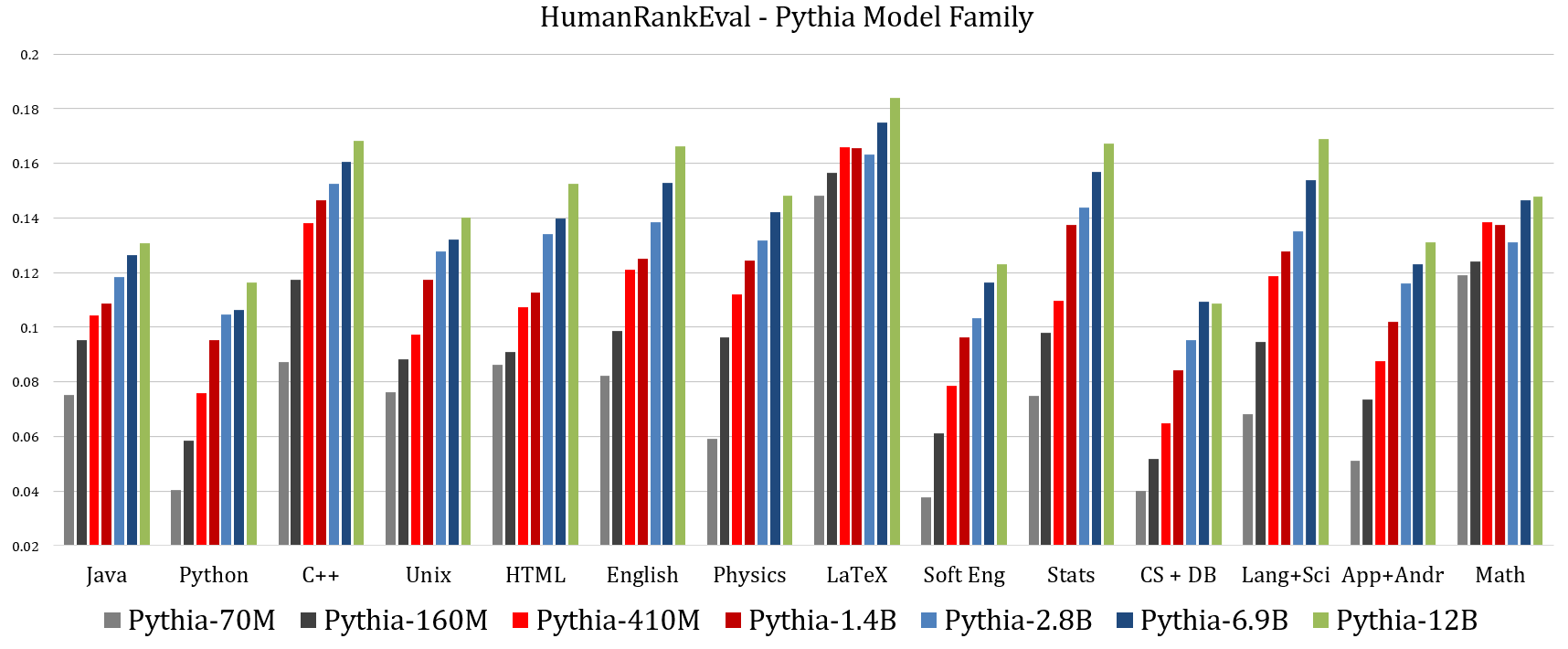}
\caption{HumanRankEval (per-topic) scores for Pythia LMs.}
\label{fig:pythia}
\end{figure*}

\subsection{Data Filtering}
\label{sec:filtering}

HRE includes QA pairs that meet the following criteria: i) the question has at least 4 answers (keep the top 4) to ensure a meaningful ranking, ii) the answers are scored by at least 40 people (10 per answer, on average) to ensure a minimum annotator pool size for each question thus giving a more reliable agreement on the rankings, iii) each answer has at least 5 votes to ensure a minimum annotator pool for each answer hence avoiding low quality responses, iv) the maximum length of each QA pair is 4,000 characters to evaluate models with shorter context windows without truncation, v) answers with identical votes are discarded (we keep the first answer with N votes) and vi) duplicate QA pairs are discarded to ensure unique QA pairs for each topic. This resulted in 7K questions (28K answers) spanning 14 topics, shown in Figure \ref{fig:stats}. The QA pairs collectively received over 700k votes (7k questions, 100+ votes per question on average) from more than 100K domain experts and enthusiasts, assuming a $\sim$20\% proportion of unique users, as in LMSYS-Chat-1M \cite{zheng2023lmsys}.

\subsection{HumanRankEval Score}
\label{sec:hre_score}

The HumanRankEval metric is based on the assumption that an LM's conversational quality can be estimated by whether the sequences it produces more frequently are more preferable to humans than the infrequent ones. 
Sequence generation tasks such as WMT \cite{barrault-etal-2020-findings}, HumanEval \cite{chen2021evaluating} and GSM8K \cite{cobbe2021training} provide the LM with a prompt (e.g. problem description), generate the output token-by-token, possibly extract the answer from the returned text, then compute the score. Alternatively, we can provide the questions as prompts to the LM, and assuming direct access to the logits of the LM being evaluated, determine the log-likelihood of the HRE human-authored answers under that model's distribution. More formally, we compute the log-likelihood of answer tokens $T_a$ using model $p$ (normalised by character length $C_a$) conditioned on the question, to obtain log-likelihoods $ll$ for each answer $a \in A$, as shown in Equation \ref{eq}.

\begin{equation}
    ll = \left[ \frac{1}{C_a} \sum \log(\frac{e^{p(t)}}{\sum_{t=1}^{T_a} e^{p(t)}}) \right] \forall a \in A 
\label{eq}
\end{equation}

\noindent Note that $C_a$ and $T_a$ are obtained only from answer tokens, a standard implementation.\footnote{We follow Eleuther's tokenizer-agnostic method of character (rather than token) length normalisation.} Subsequently, the log-likelihoods $ll$ are correlated with human rankings using Pearson \cite{freedman2007statistics} correlation. A discussion about the reasons for choosing Pearson over Spearman Rank \cite{zar2005spearman} coefficient follows in section \ref{sec:correlations}. Finally, the correlation coefficients are micro-averaged across all 7K questions to compute the HumanRankEval score. 

\begin{figure}[t]
\includegraphics[width=0.95\linewidth]{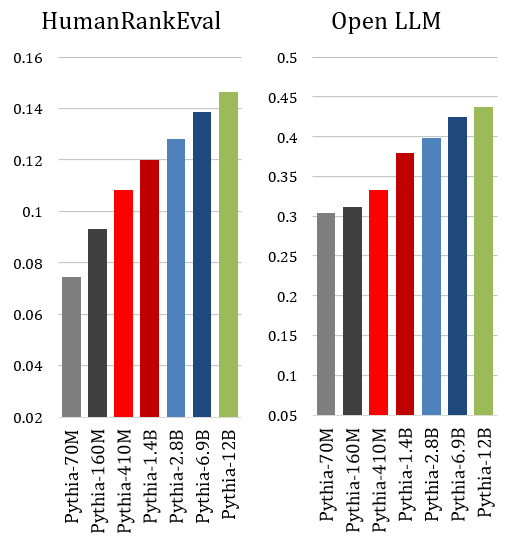}
\caption{HumanRankEval (avg) scores for Pythia LMs.}
\label{fig:pythia_overall}
\end{figure}

\begin{figure*}[t]
\includegraphics[width=\linewidth]{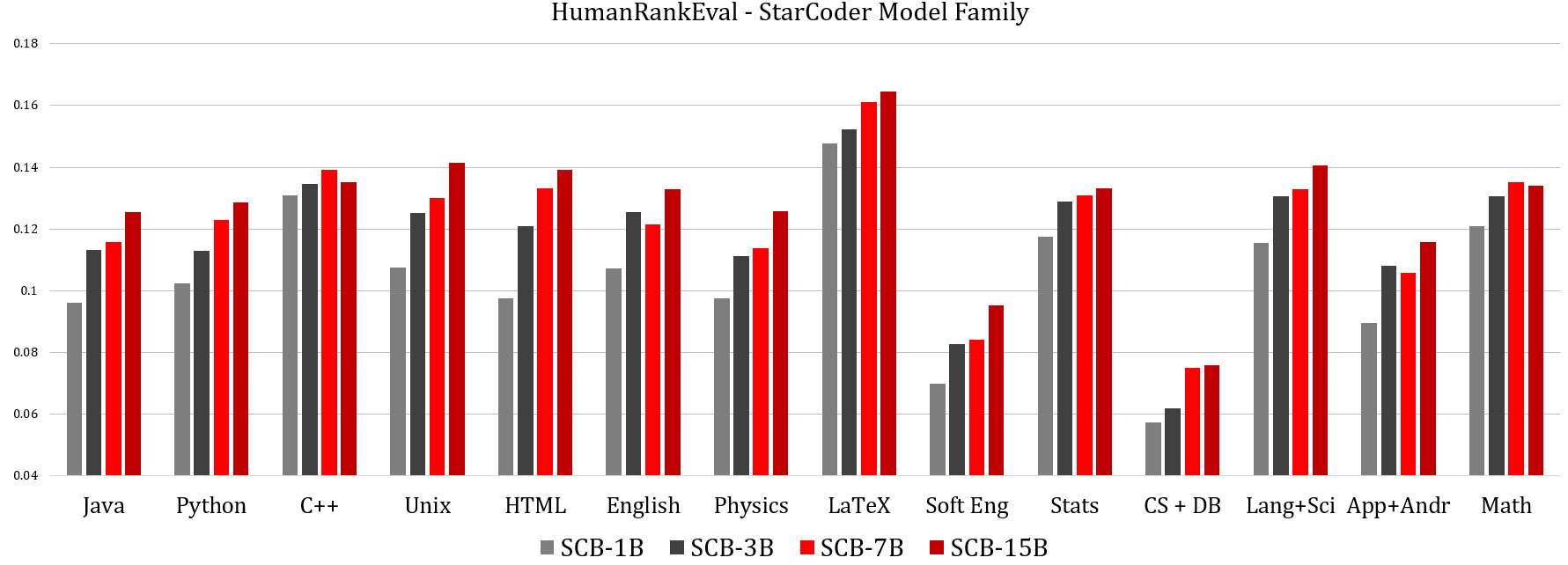}
\caption{HumanRankEval (per-topic) scores for the StarCoderBase (SCB) LMs.}
\label{fig:starcoder}
\end{figure*}

\begin{figure}[t]
\includegraphics[width=0.8\linewidth]{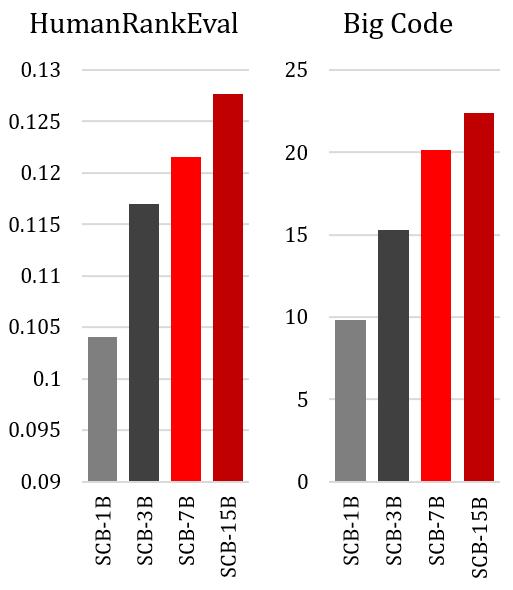}
\caption{HumanRankEval (avg) scores for the StarCoderBase (SCB) LMs versus Big Code Leaderboard.}
\label{fig:starcoder_overall}
\end{figure}

\section{Results}
\label{sec:results}

\subsection{Experimental Settings}
\label{sec:experimental}

We benchmark a broad selection of open-source LMs (pretrained and instruction-tuned) available from the Huggingface repository \cite{wolf2019huggingface}. LMs with AutoModel\footnote{\url{https://huggingface.co/docs/transformers/model_doc/auto}} and deepspeed inference\footnote{\url{https://www.deepspeed.ai/inference/}} support (tensor parallel), LM-Eval harness \cite{eval-harness} compatibility, up to 16B parameters in size were selected for efficient iteration and accessible research. This includes some of the most popular and frequently used LMs such as Llama2, Llama2-Chat \cite{touvron2023llama} (7B + 13B), CodeLlama, CodeLlama-Instruct \cite{roziere2023code} (7B + 13B), Palmyra \cite{Palmyra} and Camel \cite{Camel} (5B each, Camel is instruction-tuned), Pythia-Instruct (1.4B) from LambdaLabs\footnote{\url{https://huggingface.co/lambdalabs}}, Vicuna (7B + 13B, both instruction-tuned) from LMSYS\footnote{\url{https://hf.co/lmsys/vicuna-13b-v1.5}}, four StarCoder \cite{li2023starcoder} and seven Pythia \cite{biderman2023pythia} models (from 70M to 15.5B parameters), MPT-Chat (7B) \cite{MosaicML2023Introducing}, Zephyr (7B, instruction-tuned) Alpha + Beta \cite{tunstall2023zephyr}, WizardLM \cite{xu2023wizardlm} (13B) and Koala (13B) \cite{geng2023koala}, both instruction-tuned. Proprietary LMs were excluded as HRE needs access to the logits to compute scores.

\subsection{Increasing Model Sizes}
\label{sec:increasing}

In this section, we verify the consistency of HRE scores by observing how they increase as the size of pretrained models (code and natural language) from the same families increases. This expectation is based on the assumption that the learning capacity and general capabilities of LMs increase with the number of trainable parameters (keeping the data constant), and is supported by their performance in OpenLLM. Figures \ref{fig:pythia} and \ref{fig:starcoder} show the per-topic scores while Figures \ref{fig:pythia_overall} and \ref{fig:starcoder_overall} show the overall (micro-average) scores for seven Pythia models (70M - 12B) and four StarCoderBase models (1B - 15.5B), respectively. The Pythia models were specifically trained to study LM behavior across different sizes. As expected, different models are cleanly separated by HumanRankEval. Using a single factor ANOVA, the differences were significant between the Pythia (p=8.32-e12) and the StarCoderBase models (p=0.048).

\begin{figure*}[t]
\includegraphics[width=\linewidth]{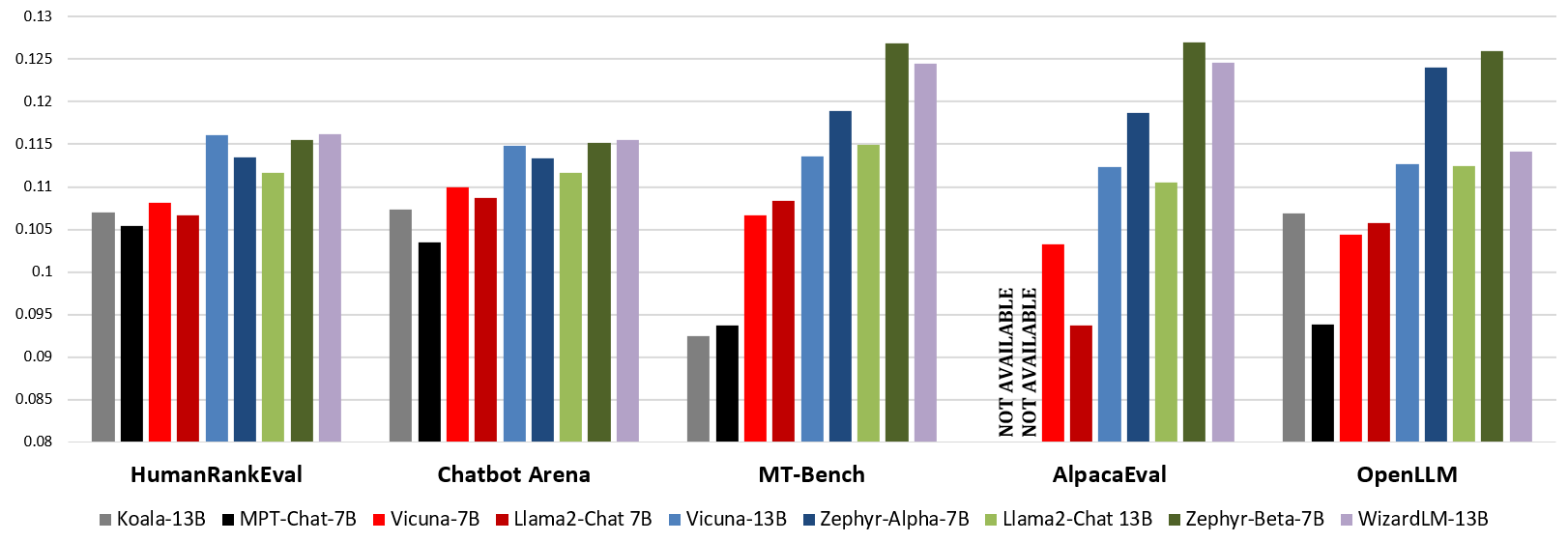}
\caption{LM rankings (normalised scores) by HumanRankEval, Chatbot Arena, AlpacaEval, MT-Bench and OpenLLM. Koala-13B and MPT-Chat-7B were not available on the AlpacaEval leaderboard at the time of writing.}
\label{fig:chatarena}
\end{figure*}

\subsection{Correlation with Human Evaluation}
\label{sec:chatbot}

\begin{figure}[h]
\centering
\floatbox[{\capbeside\thisfloatsetup{capbesideposition={left},capbesidewidth=3.5cm}}]{figure}[\FBwidth]
{\includegraphics[width=0.95\linewidth]{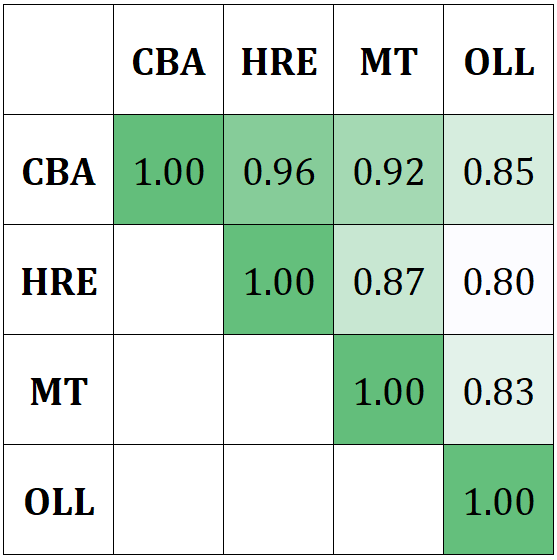}}
{\caption{Pearson correlations between HRE, OpenLLM (OLL), MT-Bench (MT) and Chatbot Arena (CBA) model rankings. MT-Bench and OpenLLM have the lowest average agreements.}
\label{fig:corr}}
\end{figure}

In order to support HRE as a reliable proxy for human judgement, we show how its scores correlate with the human-obtained Chatbot Arena ratings. To this end, in Figure \ref{fig:chatarena}, we plot the scores for various instruction-tuned models of different sizes and families. We use the latest Chatbot Arena ratings (as of writing this; 1st Nov. 2023) that were computed from $\sim$90k user votes. We observe that HRE and Chatbot Arena rankings are the most similar, while there is an obvious misalignment between rankings produced by other popular leaderboards, i.e. MT-Bench (LM-as-a-judge), AlpacaEval (LM-as-a-judge) and OpenLLM (knowledge-based). Figure \ref{fig:corr} shows the Pearson correlations between the various rankings. HRE shows the best correlation (0.96) with the human judgements of Chatbot Arena across existing leaderboards, which is to be expected as it was specifically designed for evaluating LMs as conversational assistants. OpenLLM correlates the least (0.85) with human ratings, perhaps unsurprisingly as it consists of knowledge-based automatic tasks. MT-bench's correlation (0.92) indicates that using LM-as-a-judge does offer estimations closer to human judgement than knowledge-based automatic tasks. AlpacaEval is excluded from Figure \ref{fig:corr} as rankings were unavailable for some models. 

\begin{figure}[h]
\includegraphics[width=\linewidth]{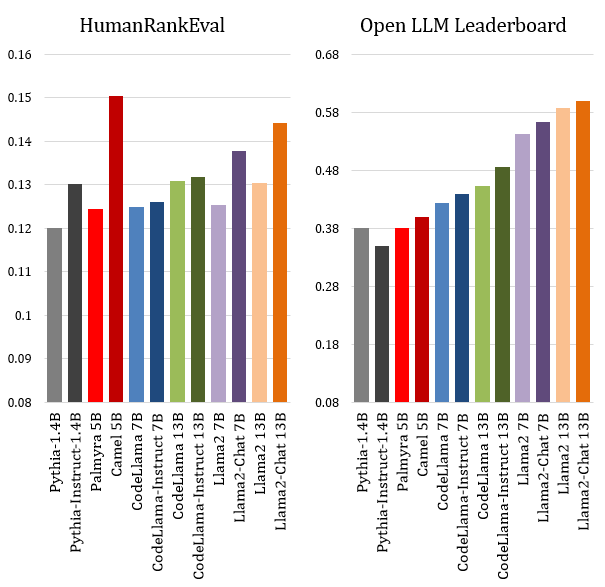}
\caption{HumanRankEval and OpenLLM scores for a selection of pretrained and instruction-tuned LMs.}
\label{fig:instruction_tuning}
\end{figure}
\vspace{-5mm}

\subsection{Instruction Tuning}
\label{sec:intruction}

HRE was developed specifically for evaluating LMs as conversational assistants, i.e. to assess the benefits of methods like instruction tuning and/or preference optimization. MT-Bench and AlpacaEval benchmark only instruction-tuned LMs thus we cannot observe how sensitive they may be to differences between pretrained and instruction-tuned models. As OpenLLM includes both types, we plot OpenLLM and HRE rankings for popular pretrained LMs and their instruction-tuned variants in Figure \ref{fig:instruction_tuning}. We observe a divergence between the two leaderboard rankings, further indicated by the low correlation between them (0.3, Pearson). OpenLLM scores underestimate the impact of instruction tuning and/or preference optimization on models' ability to follow human instructions. This is not surprising since most of its tasks assess specific types of knowledge, but this is not necessarily indicative of how an LM would perform these tasks in an open-ended conversational setting. On the other hand, we can see that Camel-5B shows a large improvement in HRE after instruction-tuning compared to its base model, Palmyra-5B (no contamination suspected, see Section \ref{sec:contamination}). Similarly, Llama2-13B obtains a higher score on the OpenLLM leaderboard than Llama2-Chat-7B, however, HRE is able to correctly detect the superior instruction-following ability of the smaller model, confirmed by the Chatbot Arena ratings in Figure \ref{fig:chatarena}. Another example (from same figure) shows an equal or higher preference by humans for Vicuna (13B) over Zephyr (7B) models despite the latter showing a significantly higher score on the OpenLLM Leaderboard. Overall, LMs fine-tuned with instruction data tend to show a noticeable improvement in HRE scores over their "vanilla" pretrained counterparts, however, there are exceptions. We hypothesise that including Self-Instruct \cite{wang2022self} data (LM-generated, automatically filtered outputs used for fine-tuning code LMs) in CodeLlama-Instruct training may be causing the weak improvement as \citet{wang2023far} have shown that training with such data adversely affected performance across factual, multilingual and reasoning tasks. 

\section{Discussion}

\subsection{Pearson over Spearman}
\label{sec:correlations}

The human and LM scores over which we calculate correlation are continuous rather than monotonically distinct. This indicates that the Pearson coefficient could be more fitting for HRE over Spearman.
This is further supported by observing that correlating the likelihoods/votes themselves or the derived rankings, results in higher agreement with human ratings in Chatbot Arena (see Figure~\ref{fig:corr}). Correlation would be lower for Spearman (0.85) than Pearson (0.96), suggesting a better fit for the latter metric. Empirically, we also observed that using Pearson results in a clearer separation of models. Figures~\ref{fig:pythia} and \ref{fig:starcoder} show the individual topic scores while Figures~\ref{fig:pythia_overall} and \ref{fig:starcoder_overall} show the averages over 14 topics. We can observe that Pearson correlation monotonically increases for 11 out of 14 topics for Pythia models and 10 out of 14 for the StarCoderBase models. On the other hand, Spearman correlation leads to a less clear separation of models, with only 5 out of 14 topics showing a monotonic increase for Pythia and 3 out of 14 for StarCoderBase respectively ( Figures~\ref{fig:appendix_pythia} and~\ref{fig:starcoder_spearman} in the Appendix).

\subsection{Data Contamination}
\label{sec:contamination}

Training LMs on content sourced from StackOverflow and/or StackExchange is not uncommon, e.g. the training data of reward models for Llama2-Chat includes StackExchange data while 2\% of Llama1 \cite{touvron2023llama} pretraining data comes from StackExchange. We posit that instruction-tuning on \textit{QA pairs} that overlap with HRE would be the most likely cause of overestimated scores, rather than pretraining on raw web pages. HumanRankEval's high correlation with human ratings (see Figure~\ref{fig:corr}) indicates low data contamination in the models we examined. We posit that unless we assume an almost uniform data contamination, we would have observed some inflated scores for some models, leading to lower correlation. According to their model cards, the benchmarked LMs such as Camel, Vicuna, Koala, MPT-Chat, Pythia-Instruct, Zephyr and Llama-Chat were not instruction-tuned with our data sources yet they show a strong improvement in HRE scores. However, not all LMs provide detailed training information hence the risk of contamination would in those cases be difficult to determine. The most appropriate future-proof action may be deduplicating training data against HumanRankEval to mitigate risks of contamination and accidental score inflation.

\section{Conclusions}
\label{sec:conclusion}

Multiple benchmarks have been proposed for evaluation of LMs as conversational assistants. However, these are either not specifically designed for this purpose, rely on large (usually proprietary) LMs as the ground truth, or are difficult to scale in terms of sourcing reliable human judges. We have therefore introduced HumanRankEval, a novel automatic evaluation task that comprises a dataset of human-authored questions and answers coupled with a metric. The votes for each question were obtained from over 100 participating domain experts (on average), resulting in high-quality human preferences. HRE performs evaluation by measuring how well the LM's ``preferences'', estimated as log-likelihoods of answers, correlate with human ratings. To validate HRE, we demonstrated that it cleanly separates pretrained and instruction-tuned LMs of various sizes, and showed that its scores correlate well with human ratings. Relative to knowledge-based evaluation, HRE is particularly adept at detecting changes to LMs' behaviour introduced by instruction-tuning and/or preference optimization. While knowledge-based automatic evaluation can test for specific skills, undesirable biases and essential world knowledge, we expect HRE to accelerate the development of LMs as conversational assistants by providing unique insights.

\section{Limitations}
\label{sec:limits}

Human preferences for our purposes were treated as a composite attribute, and no individual components such as helpfulness, factual correctness, timeliness, safety and so on can be estimated individually by HumanRankEval. LMs scoring higher on HRE are not necessarily more factually correct, less biased or more safe hence researchers are advised to conduct separate evaluation(s) to explicitly test for such behaviours. We acknowledge that, unlike knowledge-based evaluation, the ground truth of human preferences cannot be obtained with the same level of exactitude. HumanRankEval is a new addition to the current consensus and it is possible that the ground truth of human preferences may not be adequately described by any single metric or benchmark. While HRE covers a diverse collection of topics, there are specialist domains that may not be included, but are desired by some researchers. In those cases, we recommend to follow our methodology to extend HRE coverage to new domains that may be of interest. This applies to additional languages as HumanRankEval is overwhelmingly composed of English language content. Neither the StackOverflow nor the StackExchange data have specified any licence information, instructions for intended use or the presence of undesirable content. We subsample the data as is, relying on the corresponding creators of the archives for following appropriate steps. Lastly, we advise that researchers do not solely rely on HRE to verify that a model can be released for public use, and we recommend that human judgement is consulted instead. 

\section*{Acknowledgements}

We want to thank the MindSpore team members for the technical support.\footnote{\url{https://github.com/mindspore-ai}}\footnote{\url{https://mindspore.cn/}}

\bibliography{custom}
\appendix

\section{Appendix}
\label{sec:appendix}

\begin{figure*}
\includegraphics[width=0.8\linewidth]{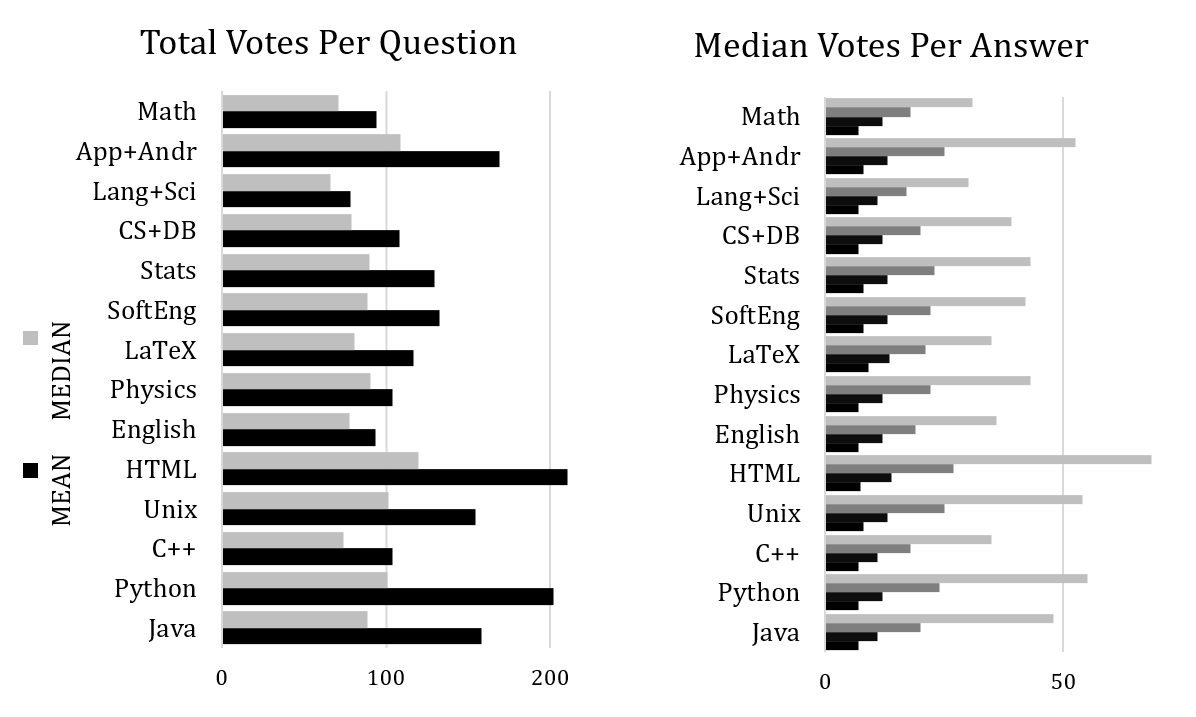}
\caption{Total votes received per question (median, mean) and median votes received per answer.}
\label{fig:appendix_votes}
\end{figure*}

\begin{figure*}
\includegraphics[width=\linewidth]{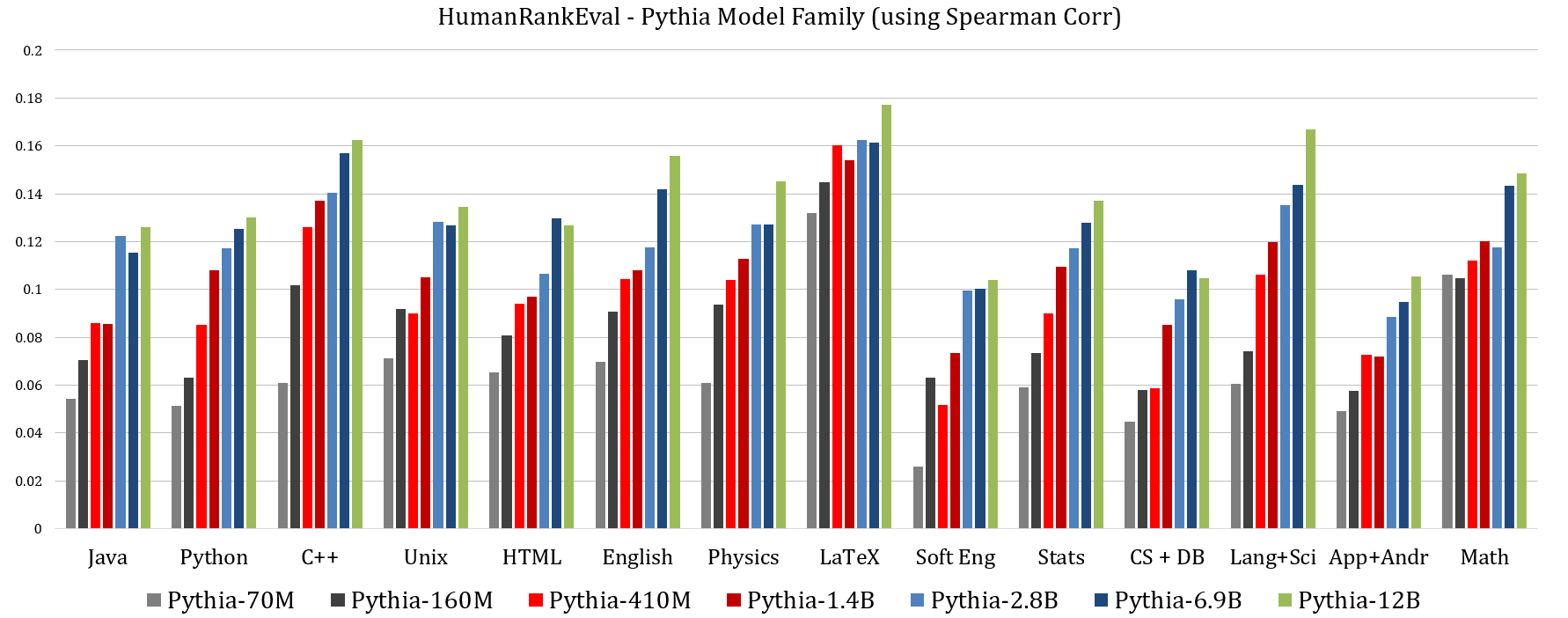}
\caption{HRE using Spearman correlation (per-topic and overall scores) for the Pythia LMs.}
\label{fig:appendix_pythia}
\end{figure*}

\begin{figure*}
\includegraphics[width=\linewidth]{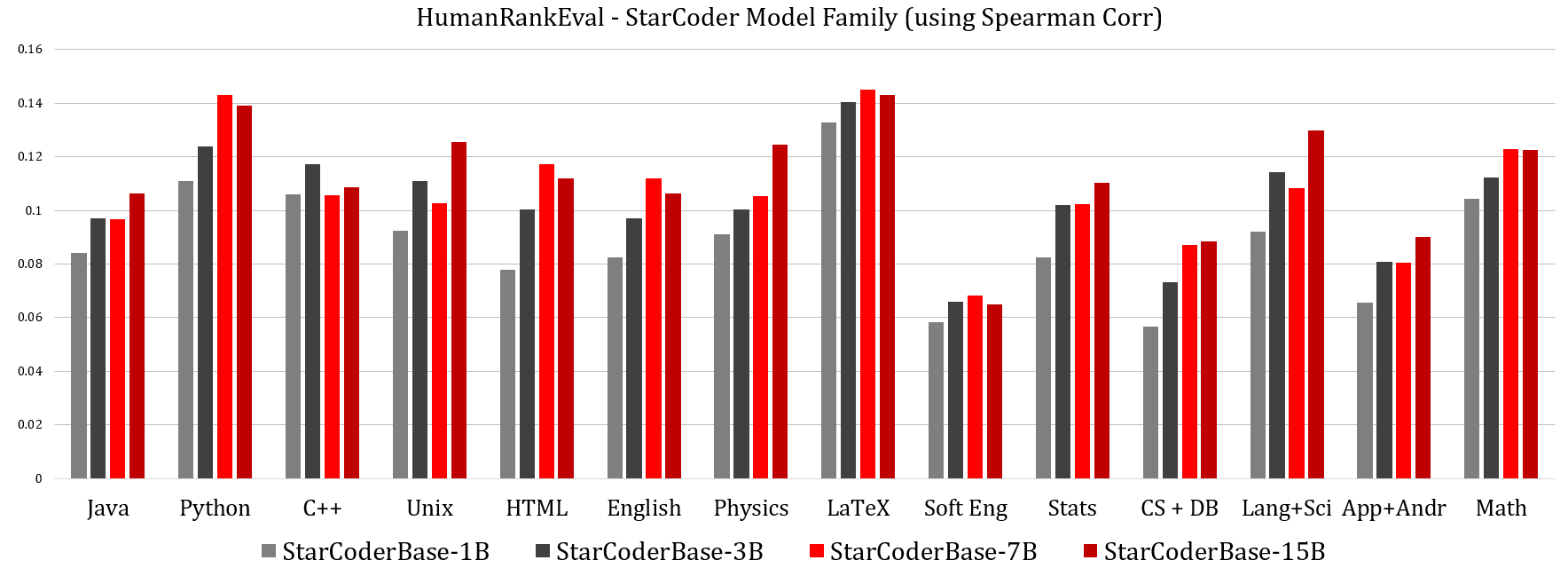}
\caption{HRE using Spearman correlation (per-topic and overall scores) for the StarCoderBase LMs.}
\label{fig:starcoder_spearman}
\end{figure*}

\end{document}